\documentclass[letterpaper, 10 pt, conference]{ieeeconf}  % Comment this line out if you need a4paper
\IEEEoverridecommandlockouts                              % This command is only needed if 
\usepackage{graphics} % for pdf, bitmapped graphics files
\usepackage{epsfig} % for postscript graphics files
\usepackage{mathptmx} % assumes new font selection scheme installed
\usepackage{times} % assumes new font selection scheme installed
\usepackage{amsmath} 
\usepackage{amssymb}  
\makeatletter
\let\NAT@parse\undefined
\makeatother

\usepackage{cite}

\usepackage{hyperref}
\hypersetup{
    colorlinks=true,
    citecolor=black,
    linkcolor=black,
    urlcolor=blue
}

\usepackage{mathtools}
\usepackage{cleveref}
\crefname{figure}{Fig.}{Figs.}
\Crefname{figure}{Figure}{Figures}

\usepackage{graphicx}

\usepackage[font=footnotesize]{caption}

\usepackage{wrapfig}
\usepackage{stfloats}
\usepackage{gensymb}
\usepackage{array}
\usepackage{booktabs}
\usepackage{multirow}

\title{\LARGE \bf
Extending the Law of Intersegmental Coordination: Implications for Powered Prosthetic Controls} 
\author{Elad Siman Tov$^{1}$ and Nili E. Krausz$^{2}$
\thanks{*This work was funded by the Israel Science Foundation grant 2937/24}
\thanks{$^{1}$Elad Siman Tov is with the Faculty of Mechanical Engineering,
        Technion – Israel Institute of Technology, 32000 Haifa, Israel
        {\tt\small elad.sim@campus.technion.ac.il}}%
\thanks{$^{2}$Nili E. Krausz is with the Faculty of Mechanical Engineering,
        Technion – Israel Institute of Technology, 32000 Haifa, Israel
        {\tt\small nili.krausz@technion.ac.il}}%
}

\begin{document}

\maketitle
\thispagestyle{empty}
\pagestyle{empty}
\begin{abstract}
Powered prostheses are capable of providing net positive work to amputees and have advanced in the past two decades. However, reducing amputee metabolic cost of walking remains an open problem. The Law of Intersegmental Coordination (ISC) has been observed across gaits and previously implicated in energy expenditure of walking, yet it has rarely been analyzed or applied within the context of lower-limb amputee gait. 
This law states that the elevation angles of the thigh, shank and foot over the gait cycle covary.
In this work, we developed a method to analyze intersegmental coordination for lower-limb 3D kinematic data, to simplify ISC analysis. Moreover, inspired by motor control, biomechanics and robotics literature, we used our method to extend ISC to a new law of coordination of moments. We find these Elevation Space Moments (ESM), and present results showing a moment-based coordination for able bodied gait. 
We also analyzed ISC for amputee gait with powered and passive prostheses, and found that while elevation angles remained planar, the ESM lacked planar coordination. We present an ISC-driven powered prosthetic control framework, using healthy coordination as a constraint to predict the shank angles/moments to compensate for alterations due to a passive foot. We developed the ISC3d toolbox that is freely available online, which may be used to compute kinematic and kinetic ISC in 3D. This provides a means to further study the role of coordination in gait and may help address fundamental questions of the neural control of human movement.

\end{abstract}
%%%%%%%%%%%%%%%%%%%%%%%%%%%%%%%%%%%%%%%%%%%%%%%%%%%%%%%%%%%%%%%%%%%%%%%%%%%%%
\section{Introduction}

Transfemoral amputees have been shown to have elevated metabolic cost when walking compared to able-bodied gait \cite{van_schaik_metabolic_2019} as well as commonly exhibiting prominent hip compensations, such as circumduction \cite{gailey_review_2008}. Typically these individuals are clinically prescribed a microprocessor knee (MPK) and a passive prosthetic foot (e.g. carbon-fiber dynamic response feet), which are often separate commercial products targeting individual joints. These devices dissipate energy and cannot provide positive power which might explain elevated metabolic cost for amputees \cite{winter_biomechanics_1991}. In the past several decades researchers have developed powered prostheses \cite{sup_design_2008,tran_lightweight_2022} capable of providing net positive work to the amputee, with the goal of reducing metabolic cost \cite{au_powered_2009}. 

Above-knee powered prostheses have typically used joint level impedance control at the knee and ankle \cite{sup_design_2008}. Inspired by \cite{hogan_impedance_1985}, for each desired gait behavior the torque-angle relationship at each joint is parameterized over discrete phases of the gait cycle. Recently, state-of-the-art powered prosthesis controllers expand on this approach by estimating a continuous gait \textit{phase variable} to parameterize their control law for the knee and ankle joints over the gait cycle \cite{rachel_gehlhar_review_2023}. This control method can enable a wide variety of gait behaviors; however, there remain existing limitations with this approach. For example, understanding how changes in knee or ankle parameters affect energetic cost or induce unwanted hip compensations is challenging.

An important result with implications for reducing metabolic cost in amputee gait was presented almost three decades ago \cite{luigi_bianchi_kinematic_1998}. Researchers showed a relationship between the mechanical energy of walking and the kinematic coordination of the human leg in elevation angles (segment orientations as measured from the global vertical axis). While it is standard to present kinematic data in joint angles, elevation angles have been observed to follow the so-called Law of Intersegmental Coordination (ISC) \cite{borghese_kinematic_1996}. The ISC phenomenon has been observed across gaits, speeds, and inclines \cite{luigi_bianchi_kinematic_1998,borghese_kinematic_1996,ivanenko_modular_2007,ivanenko_origin_2008,dewolf_kinematic_2018,israeli-korn_intersegmental_2019} whereby the elevation angles of the thigh, shank, and foot segments (when measured in the sagittal plane) form trajectories that covary along a plane (termed the Covariation Plane or CVP). The tilt of the CVP was found by \cite{luigi_bianchi_kinematic_1998} to be tied to the mechanical energy cost at a wide range of gait speeds. Although ISC is a kinematic determinant of human locomotion \cite{borghese_kinematic_1996}, the study of ISC in the context of amputee gait has been limited \cite{krausz_asymmetric_2023}.

There are clear implications of ISC for powered prosthetics. Due to the phase-based parameterization, existing controllers used by ankle-knee prostheses implicitly impose a \textit{coupling} between joints, whether in joint angles, joint torques, or even joint level impedance. Even in a decoupled control architecture, when parameterizing both joint-level knee and ankle controllers by a single phase variable their references do not change independently. This approach originated from bipedal robot control methods (see \cite{westervelt_feedback_2018}), and has been implemented in experimental powered prosthetic devices for multiple locomotion tasks \cite{gregg_virtual_2014,t_kevin_best_phase-variable_2021,t_kevin_best_data-driven_2023,cortino_data-driven_2024,gehlhar_control_2022}. Alternatively, in \cite{sullivan_unified_2025} an adaptive-control based method was presented that explicitly couples the joint controllers (without a phase variable) with position and torque synergies between joints, and in \cite{kellyTaskSpaceControl2025} a task space controller was demonstrated to coordinate a user with a powered ankle prosthesis. Thus, existing controllers for powered prostheses aim at achieving coordinated gait, without explicitly considering ISC.

In this work, we analyze ISC in amputee gait with powered prostheses and compare to behavior of able-bodied individuals. We present a new method for transforming reported joint angles into elevation angles, which can then be used to assess ISC. We then consider how ISC could be extended to dynamics. Joint moments in the sagittal plane have previously been shown to be related to one another by the \textit{Total Support Moments} (see \cite{winter_overall_1980,hof_interpretation_2000}). Winter claimed this support moment suggests a compensatory mechanism where joint moments compensate for one another. Inspired by this work and ISC, we hypothesize that joint moments in a transformed elevation angle space will also covary as do elevation angles. Here we evaluate the potential coordination of \textit{elevation space moments} ESM to assess whether dimensionality reduction exists for moments as well as kinematics in elevation angle coordinates and present a novel approach for finding (ESM). We also consider how this dimensionality reduction may be altered in amputee gait. Our conjecture, is that a deeper understanding of these phenomena could lead to a unified theory of dynamic coordination with potential as a new paradigm for control of powered prosthetic legs.
%%%%%%%%%%%%%%%%%%%%%%%%%%%%%%%%%%%%%%%%%%%%%%%%%%%%%%%%%%%%%%%%%%%%%%%%%%%%%
\section{Methods}
%%%%%%%%%%%%%%%%%%%%%%%%%%%%%%%%%%%%%%%%%%%%%%%%%%%%%%%%%%%%%%%%%%%%%%%%%%%%%

\subsection{Datasets}
Two datasets were used. An open dataset of 10 Able Bodied (AB) individuals including walking, running and stair ascent/descent activities in variable inclines and speeds \cite{reznick_lower-limb_2021}; and a dataset of 3 Transfemoral Amputees (TFA) walking at variable speeds, with their prescribed passive prosthesis and a powered prosthesis \cite{elery_effects_2020}.
Both datasets were recorded using Vicon motion capture system accompanied by an instrumented treadmill (Bertec).
In this work, for both datasets we analyzed steady-state walking, with zero incline over three speeds.
The AB individuals walked at speeds ($0.8[m/s]$, $1.0[m/s]$, and $1.2[m/s]$), and each of the three amputees (TFA1, TFA2, TFA3) walked at a slow ($0.9[m/s]$, $0.8[m/s]$, $0.8[m/s]$), normal ($1.1[m/s]$, $1.0[m/s]$, and $1.0[m/s]$) and a fast ($1.3[m/s]$, $1.2[m/s]$, and $1.2[m/s]$) self selected speed, respectively. Joint torques were provided by the used datasets and normalized by body-weight ($[Nm/kg]$).

\subsection{Elevation Angle Covariation Plane}\label{sec:CVP}

The elevation angles of the thigh ($\alpha_t$), shank ($\alpha_s$), and foot ($\alpha_f$) are computed using the transformation as will be described in \cref{sec:joint2elevation}. 
The Covariation Plane (CVP) is the visual representation of elevation angle coordination (see \cref{fig:2_elevation_cvp}). 
Principal Component Analysis (PCA) was performed for all trials to evaluate intersegmental coordination as in \cite{luigi_bianchi_kinematic_1998,borghese_kinematic_1996,ivanenko_modular_2007}. 
The projections of the thigh, shank, and foot trajectories onto the PC vectors are referred to as the PC scores.
The Planarity Index (PI) is defined as the variance accounted for by the first two PCs, and used to quantify the level of planarity of the CVP. Larger values imply a stronger planar dependency between variables as in \cite{israeli-korn_intersegmental_2019}. Here $\lambda_i$ is the $i^{th}$ PC variance, i.e. the $i^{th}$ eigenvalue of the covariance matrix.
\begin{equation}\label{eq:planarityIndex}
    PI=\frac{\lambda_1+\lambda_2}{\lambda_1+\lambda_2+\lambda_3}\cdot 100 \%
\end{equation}

In addition to evaluating the CVP and PI, the coupling at the shank-foot is considered and the linear portion associated with the swing phase of gait was fit with a linear model. 

\subsection{Transforming Joint Angles To Elevation Angles}\label{sec:joint2elevation}
Elevation angles have been used to analyze intersegmental coordination; however, these studies typically compute elevation angles directly from surface markers to approximate segment axes. Without using motion capture software, this creates a computational challenge, and may be inconsistent with standard definitions of anatomical joint angles and joint center estimates. Moreover, this leads to challenges in making broader comparisons about intersegmental coordination, as most human gait studies present data in terms of joint angles. 
Therefore, a general method to compute elevation angles from previously published joint angle data was needed. Our algorithm utilizes 3D anatomical angles computed by motion capture systems to estimate the elevation angles of the leg segments. This transformation between joint and elevation angles also provides an analytical method for transforming velocities and moments into an elevation coordinate frame, which will be shown in \cref{sec:elevJacobian}. 

We denote the flexion angle as \(\phi\), the adduction angle as \(\delta\), and the rotation angle as \(\rho\), with the lower script to represent the corresponding joint (i.e., knee flexion \(\phi_k\) and hip internal rotation as \(\rho_h\)).
Given the pelvis absolute angles ($\phi_p, \delta_p, \rho_p $) with respect to a subject global reference frame (following ISB recommendations \cite{wu_isb_1995}), the orientation of the pelvis segment frame ($R_p$) is found by the intrinsic multiplication of rotation matrices in the conventional order for the pelvis (Pelvic Tilt, Obliquity, Rotation). 
\begin{equation}\label{eq:pelvisRotMatrix}
    R_p=R_z(\phi_p)\, R_x(\delta_p) \, R_y(\rho_p)
\end{equation}
Similarly, the thigh ($R_t$), shank ($R_s$), and foot ($R_f$) segment frame orientations were found using the anatomical hip, knee, and ankle angles, with rotations in the conventional order (see Appendix \Cref{tab:sign_conv}).
\begin{equation}\label{eq:thighRotMatrix}
    R_t=R_p \, R_z(\phi_h) \, R_x (\delta_h) \, R_y (\rho_h)
\end{equation}
\begin{equation}\label{eq:shankRotMatrix}
    R_s = R_t \, R_z(\phi_k) \, R_x (\delta_k) \, R_y (\rho_k)
\end{equation}
\begin{equation}\label{eq:footRotMatrix}
    R_f = R_s \, R_z(\phi_a+90\degree) \, R_x (\rho_a) \, R_y (\delta_a)
\end{equation}

\begin{figure}[thpb]
    \centering
    \includegraphics[width=0.7\linewidth]{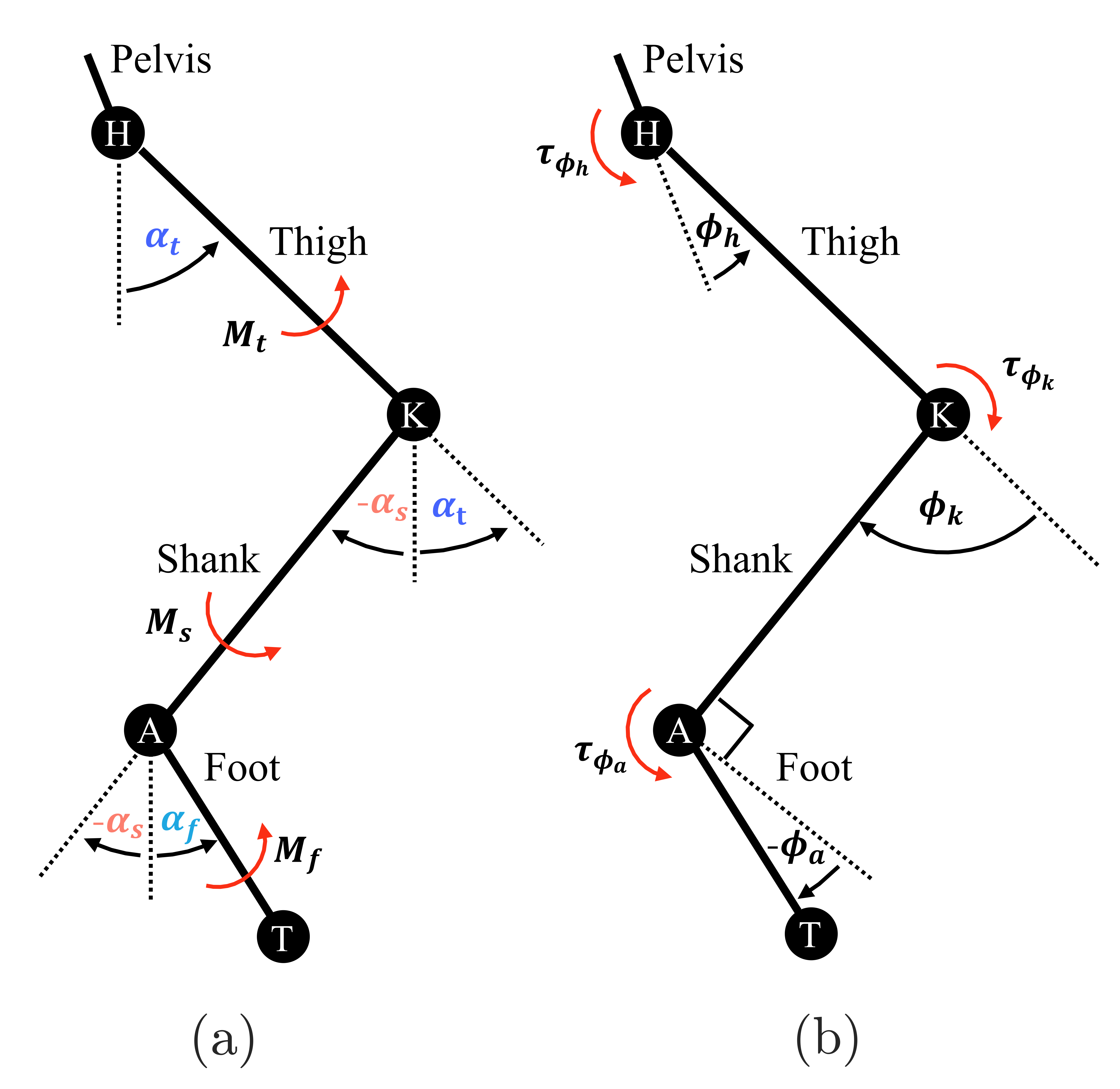}
    \caption{Schematic of (a) elevation angles ($\alpha_i$) and moments ($M_i$) in the sagittal plane, and (b) anatomical joint flexion angles ($\phi_j$) and their internal joint torques ($\tau_{\phi_j}$). Ankle dorsiflexion is assumed as positive. All moments are defined in the positive direction of their associated angle.}
    \label{fig:Joint2Elevation2D}
\end{figure}

The elevation angular velocity vector is found using the skew-symmetric matrix representation of angular velocity of each segment ($\Omega_i\in so(3)$), computed by the rotation matrices ($R_i\in SO(3)$), and their element-wise derivatives ($\dot R_i$) for a fixed frame \cite{murray_mathematical_2017}.
\begin{equation}
    \Omega_i=\dot R_iR_i^T
\end{equation}
Here the components of the angular velocity vector represented by $\Omega_i$ are the $i^{th}$ segment angular velocities in three axes (sagittal, frontal, transversal) in a subject fixed global frame.
The angular velocities of each segment $i$ are found by selecting its components from $\Omega_i$, as shown in \eqref{eq:OmegaElevationAngle}. For the left leg ($\Omega^{L}_{i}$), the rotations ($\alpha_i,\beta_i,\gamma_i$) are about (+Z,-X,-Y) axes, and for the right leg ($\Omega^{R}_{i}$), the ($\alpha_i,\beta_i,\gamma_i$) rotations are about (+Z,+X,+Y) axes in an ISB recommended frame \cite{wu_isb_1995}.
\begin{align}\label{eq:OmegaElevationAngle}
    \Omega^{R}_{i} &= \begin{bmatrix} 
        0 & -\dot{\alpha}_i & -\dot{\gamma}_i \\
        \dot{\alpha}_i & 0 & \dot{\beta}_i \\
        \dot{\gamma}_i & -\dot{\beta}_i & 0 
    \end{bmatrix} &
    \Omega^{L}_{i} &= \begin{bmatrix} 
        0 & -\dot{\alpha}_i & \dot{\gamma}_i \\
        \dot{\alpha}_i & 0 & -\dot{\beta}_i \\
        -\dot{\gamma}_i & \dot{\beta}_i & 0 
    \end{bmatrix}
\end{align}
In this work we consider only the elevation angular velocities in the sagittal plane ($\dot \alpha$) using 
\eqref{eq:alphaDot}, because coordination is observed within a subspace of $\alpha$.
\begin{equation}\label{eq:alphaDot}
    \dot \alpha =[\Omega_{p,(2,1)},\Omega_{t,(2,1)},\Omega_{s,(2,1)},\Omega_{f,(2,1)}]^T
\end{equation}
For each segment $i$ its elevation angle $\alpha_i$ is computed using the projection of the segment axis onto the sagittal plane as in \eqref{eq:calc_elevation_angles}. This is consistent with formulas presented in \cite{borghese_kinematic_1996}.
\begin{align}\label{eq:calc_elevation_angles}
    &\alpha_i = \text{atan2} (-R_{i,{(1,2)}},R_{i,(2,2)})
\end{align}

\subsection{Elevation Space Moments (ESM)}\label{sec:elevJacobian}
% ------------------------------------ % 
In robotics, a typical method to map generalized force and velocity vectors between joint space and task space uses the Jacobian and its pseudo-inverse \cite{siciliano_springer_2008}.
Inspired by this approach, we define a \textit{elevation space Jacobian} $(J\in \mathbb{R}^{12 \times 12})$, to transform anatomical joint velocities ($\dot{q}$) to segment elevation velocities ($[\dot{\alpha} ,\dot{\beta}, \dot{\gamma}]^T$)

\begin{equation}\label{eq:alphasDotEQJqdot}
    \begin{bmatrix} 
        \dot \alpha  \\
        \dot \beta  \\
        \dot \gamma  \\ 
    \end{bmatrix} = J(q) \dot q 
\end{equation}

Following this logic, anatomical joint moments ($\tau$) map to their projected moments in the elevation angle space ($M$) as shown in \eqref{eq:transformedMoments}
\begin{equation}\label{eq:transformedMoments}
    M=(J^T(q))^{\dagger} \tau
\end{equation}

\noindent where $q\in \mathbb{R}^{12}$ is the vector of 3D anatomical angles of the pelvis, hip, knee and ankle joints, and $\tau\in \mathbb{R}^{12}$ is the vector of 3D anatomical joint torques (see Appendix \ref{appendix:signCorrections}).

We derived $J(q)$ using the MATLAB Symbolic Math Toolbox. The algorithm was implemented in our ISC3d toolbox which can be freely accessed online \cite{siman_tov_isc3d_2026}. We hypothesized that coordination would also be found in elevation space moments, which we quantified using the Planarity Index as described in \eqref{eq:planarityIndex} (see  \cref{sec:CVP}).

% % ---------------------------------------------------------%
\subsection{Elevation Space Moments in 2D}\label{sec:interpESM}
 
Consider a simplified 2D case in the sagittal plane shown in \cref{fig:Joint2Elevation2D}, where out-of-plane terms $\left(R_x(\cdot)R_y(\cdot)\right)$ are ignored. Under these conditions, \eqref{eq:calc_elevation_angles} reduces to a relation between joint flexion angles ($\phi_j$) and sagittal plane elevation angles ($\alpha_i$) given in \eqref{eq:trans2Djoint2elev}  (see sign corrections in Appendix \Cref{tab:sign_conv}).
\begin{align}\label{eq:trans2Djoint2elev}
    \begin{bmatrix} 
        \alpha_p  \\
        \alpha_t  \\
        \alpha_s  \\
        \alpha_f 
    \end{bmatrix} =
    \begin{bmatrix}
        -\phi_p  \\
        -\phi_p + \phi_h  \\
        -\phi_p + \phi_h - \phi_k   \\
        -\phi_p + \phi_h - \phi_k + \phi_a + 90\degree 
    \end{bmatrix}
\end{align}
Accordingly, we obtain a transformation between joint angle and elevation angle velocities in the sagittal plane in \eqref{eq:trans2Djoint2elevVel}.
\begin{align}\label{eq:trans2Djoint2elevVel}
    \begin{bmatrix} 
        \dot \alpha_p  \\
        \dot \alpha_t  \\
        \dot \alpha_s  \\
        \dot \alpha_f 
    \end{bmatrix} =
    \begin{bmatrix}
        -\dot \phi_p \\
        -\dot \phi_p + \dot \phi_h  \\
        -\dot \phi_p + \dot \phi_h - \dot \phi_k   \\
        -\dot \phi_p + \dot \phi_h - \dot \phi_k + \dot \phi_a 
    \end{bmatrix}
\end{align}

\begin{figure}[h]
  \centering
  \includegraphics[width=0.95\linewidth]{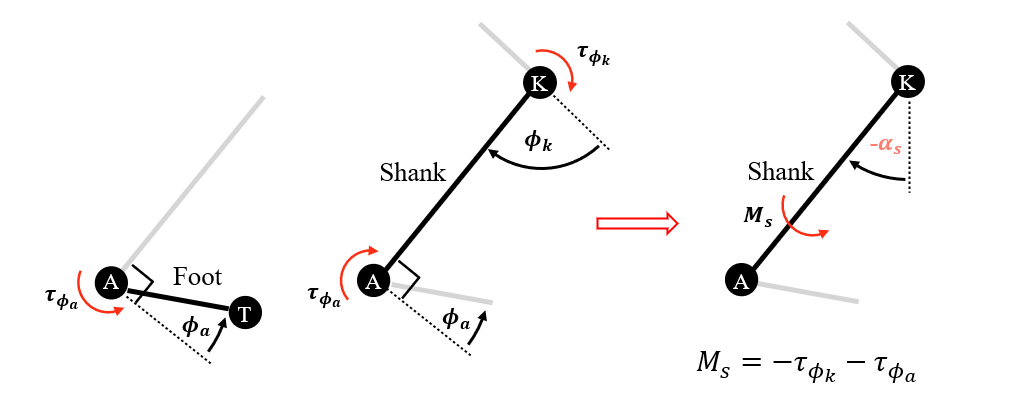}
  \caption{Projections of anatomical joint torques onto the shank segment elevation angle coordinate ($\alpha_s$) in a simplified 2D case. The coordinate ($\alpha_s$) is defined as positive in the CCW direction. The shank elevation space moment ($M_s$) consists of the knee flexion ($\tau_{\phi_k}$) torque which acts CW on the shank, and CW reaction to the ankle dorsi-flexion ($\tau_{\phi_a}$) torque, which acts CCW on the foot, as in \cref{eq:trans2DtotalPower}. 
  }
  \label{fig:interp_ESM}
\end{figure}

Because the total instantaneous external power generated ($P_T$) is invariant in the coordinates in which it is represented, $P_T$ can be given as the sum of products of internal joint torques and the relative joint velocities (defined in the same directions).
A simplified version of ESMs are found as the combined joint torque terms ($\tau_{\phi_j}$) on the right hand side of \eqref{eq:trans2DtotalPower}.
\begin{align}\label{eq:trans2DtotalPower}
    P_T=
    \begin{bmatrix} 
        \tau_{\phi_p}  \\
        \tau_{\phi_h}  \\
        \tau_{\phi_k}  \\
        \tau_{\phi_a} 
    \end{bmatrix}^T
    \begin{bmatrix} 
        \dot \phi_p  \\
        \dot \phi_h  \\
        \dot \phi_k  \\
        \dot \phi_a 
    \end{bmatrix} =
    \begin{bmatrix*}[r] 
        - \tau_{\phi_p}- \tau_{\phi_h} \\
        \tau_{\phi_h}+\tau_{\phi_k}  \\
        -\tau_{\phi_k} -\tau_{\phi_a}  \\
        \tau_{\phi_a} 
    \end{bmatrix*}^T
    \begin{bmatrix} 
        \dot \alpha_p  \\
        \dot \alpha_t  \\
        \dot \alpha_s  \\
        \dot \alpha_f 
    \end{bmatrix} 
\end{align}
These terms represent the net contributions of anatomical joint torques onto the segment elevation angle coordinates. \cref{fig:interp_ESM} provides an example interpretation for the shank ESM.

We used 3D anatomical angles ($q$) to derive the elevation space Jacobian ($J$) to account for out of plane terms; however, in this work we focus on a reduced elevation space in the sagittal plane ($\alpha$, $\dot \alpha$ and $M_\alpha$), as intersegmental coordination is traditionally viewed in terms of the thigh, shank and foot. Here we define $\dot{\alpha} = [\dot{\alpha}_t , \dot{\alpha}_s, \dot{\alpha}_f]^T$ and $M_\alpha=[M_t, M_s, M_f]^T$ as the sagittal plane elevation space velocities and moments, where we ignore pelvis terms. In future work, we will consider 3D coordination including the $\beta,\gamma$ angles.

The evaluation of sagittal plane powers given these assumptions is presented in \cref{sec:powerEvaluation}. We compare the total instantaneous power in joint space ($P_T=\tau^T\dot {q}$) and in elevation space ($P_T=M_\alpha^T\dot \alpha$).

%%%%%%%%%%%%%%%%%%%%
\subsection{The CVP as a constraint}
We consider how planar coordination could be leveraged for prosthesis control. Our conjecture is that by controlling a prosthesis to coordinate appropriately, we may reduce unwanted compensatory movements. Here we present an example case for utilizing the CVP towards control of a powered knee and passive ankle combination. 
Commercially available prosthetic legs for above-knee amputees often combine a passive foot with a powered knee/MPK. These are typically separate products designed to solve independent joint-level problems and consequently, do not actively coordinate.
We propose that imposing the CVP as a constraint on the control of a prosthesis could allow the shank to compensate for a passive prosthetic ankle-foot while attempting to achieve healthy thigh behavior.

Let the elevation angle CVP be defined by the constraint $g_\alpha(\alpha_t,\alpha_s,\alpha_f)=0$ and the elevation space moments CVP as $g_M(M_t,M_s,M_f)=0$. The thigh components are given by a mean value of the AB subject, and the foot components are taken from the TFA (Passive) foot behavior. Thus, the predicted shank angle/moment profile for coordination, is found in \eqref{eq:AngleFF} and \eqref{eq:MomentFF}, respectively.
\begin{equation}\label{eq:AngleFF}
 \alpha_s= f_\alpha(\alpha_t,\alpha_f)
\end{equation}
\begin{equation}\label{eq:MomentFF}
 M_s= f_M(M_t,M_f)
\end{equation}

\noindent This desired shank angle/moment could then be utilized for compensation in a powered knee controller. See \cref{sec:predictShankProfile} for example results from this approach.

%%%%%%%%%%%%%%%%%%%%%%%%%%%%%%%%%%%%%%%%%%%%%%%%%%%%%%%%%%%%%%%%%%%%%%%

\section{Results}
% 2_elevation_cvp
\begin{figure*}[tp]
    \centering
    \includegraphics[width=0.7\linewidth]{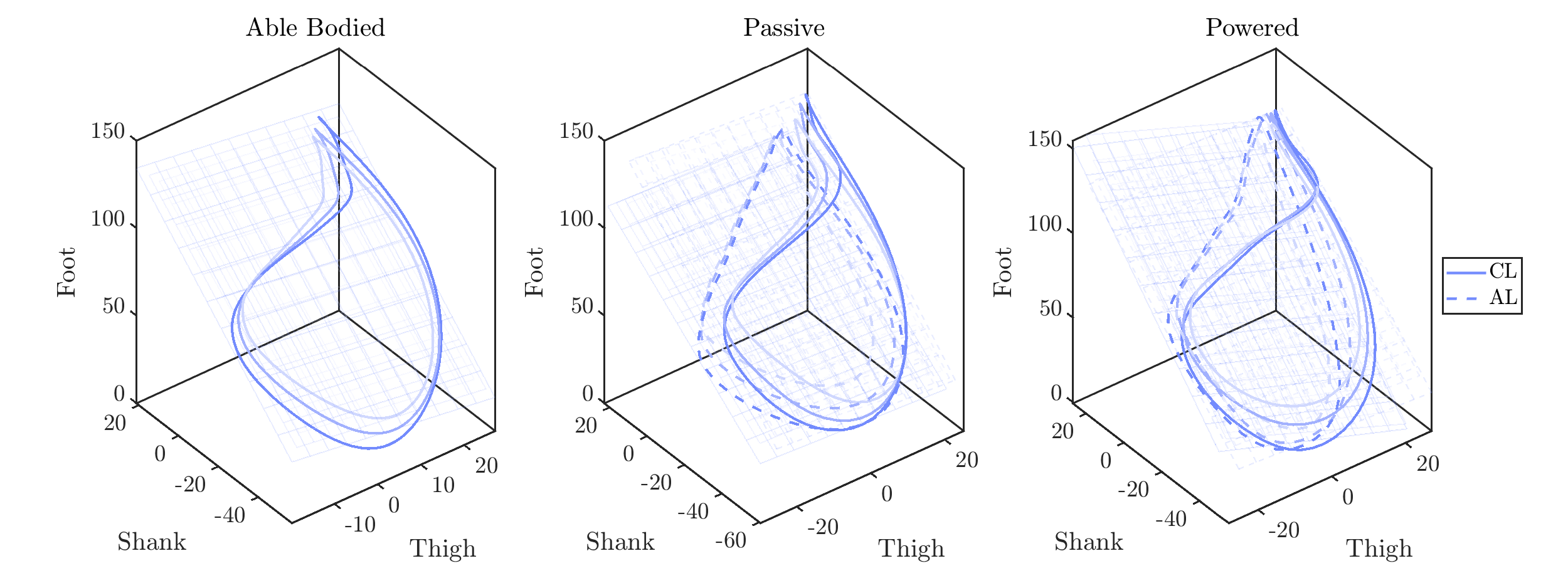}
    \caption{Elevation angles Covariation Plane. The amputated leg (AL) is indicated by a dashed line, and contralateral leg (CL) as a solid line. The plane of each loop, defined by the first and second PCs, is shown as a grid. The trajectory progresses in gait phase CCW. The top (bottom) of the loop is linked to the heel strike (toe-off) events, respectively.}
    \label{fig:2_elevation_cvp}
\end{figure*}

\subsection{Elevation Angle Covariation}
\begin{figure}[hbp]
    \centering
    \includegraphics[width=1\linewidth]{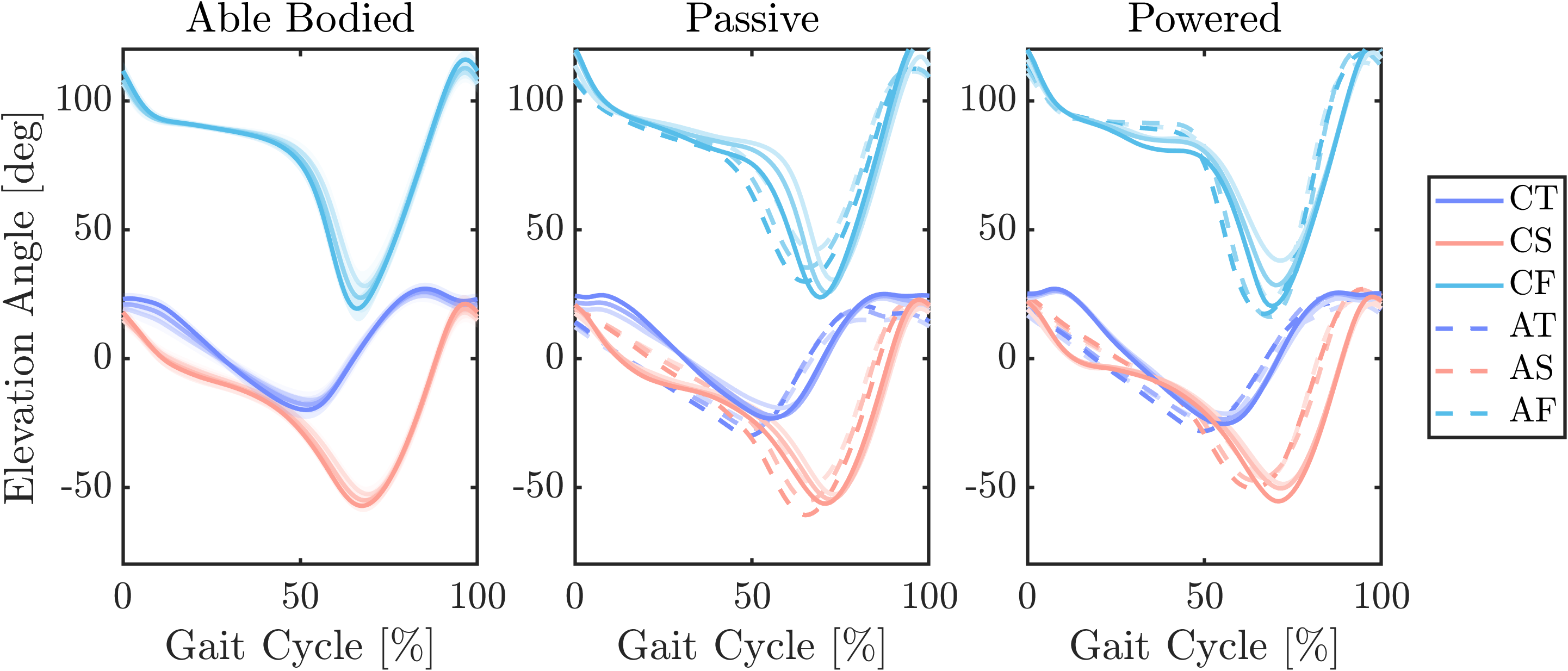}
    \caption{Elevation angles of the thigh (T), shank (S) and foot (F) computed from \eqref{eq:calc_elevation_angles}. Able bodied mean curves with shaded standard deviations (left), amputees with a passive (middle) and powered (right) prosthesis. The amputated (A) side is the dashed line, and the contralateral (C) side is the solid line. For this and \cref{fig:2_elevation_cvp,fig:3_elevation_pcs,fig:4_shank_foot,fig:6_elevation_moments,fig:7_elevation_moments_pcs,fig:mainFigure_Demonstration} means across subjects for three speeds are shown; faster speed shown in more opaque colors.}
    \label{fig:1_elevation_angles}
\end{figure}

The profiles of the mean thigh, shank, and foot elevation angles are shown across three speeds for each condition in \cref{fig:1_elevation_angles}, with asymmetries in amputee gait on both passive and powered legs. Specifically, the amputated leg has its peak elevation angle earlier in the gait cycle. The CVP results for AB and TFA subjects are given in \cref{fig:2_elevation_cvp}, with dimensionality reduction across all three conditions. Critically, the shape of the trajectory on the plane has noticeable changes during early stance for the amputated leg.
% 1_elevation_angles

\begin{figure}[h!]
    \centering
    \includegraphics[width=1\linewidth]{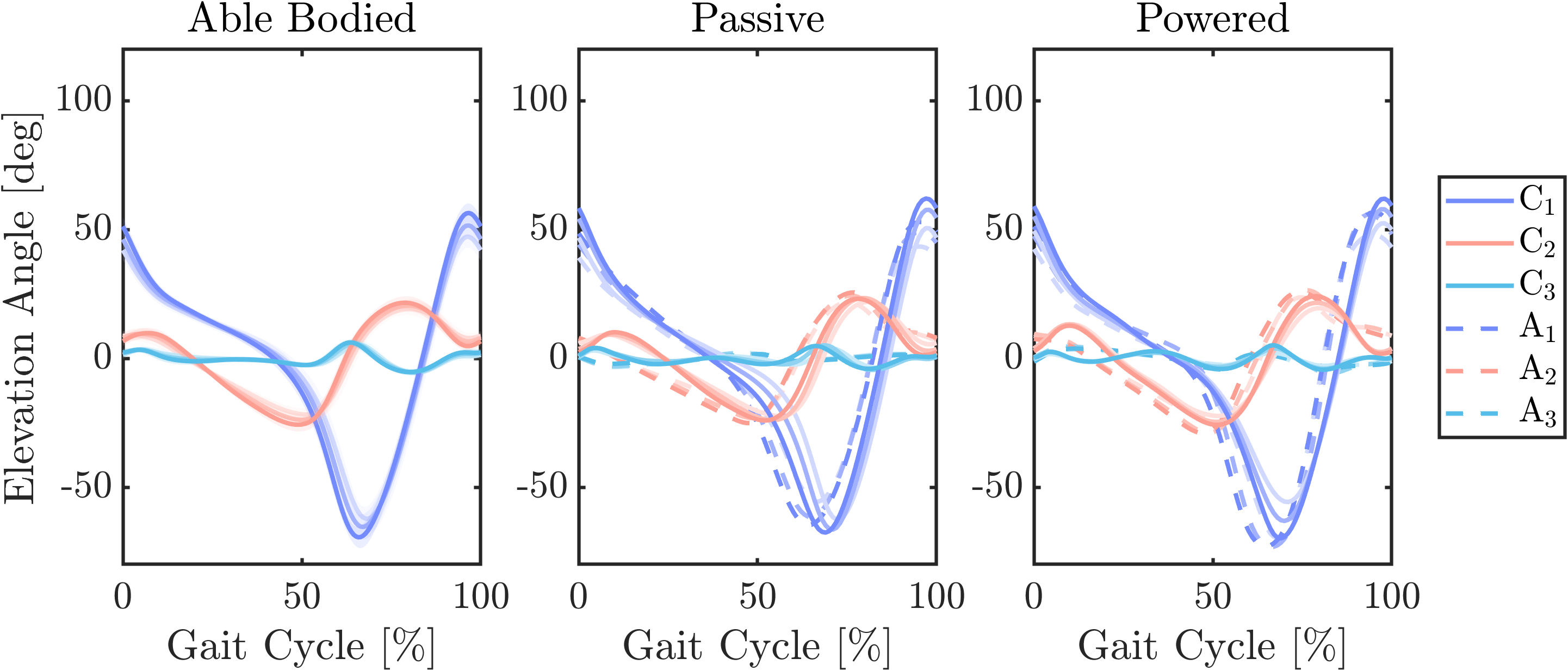}
    \caption{Principal Component scores of the elevation angles (as in \cref{fig:1_elevation_angles}), with indices ordered by decreasing PC variance for amputated (A) and contralateral (C) legs.}
    \label{fig:3_elevation_pcs}
\end{figure}

\begin{figure}[htb]
    \centering
    \includegraphics[width=1\linewidth]{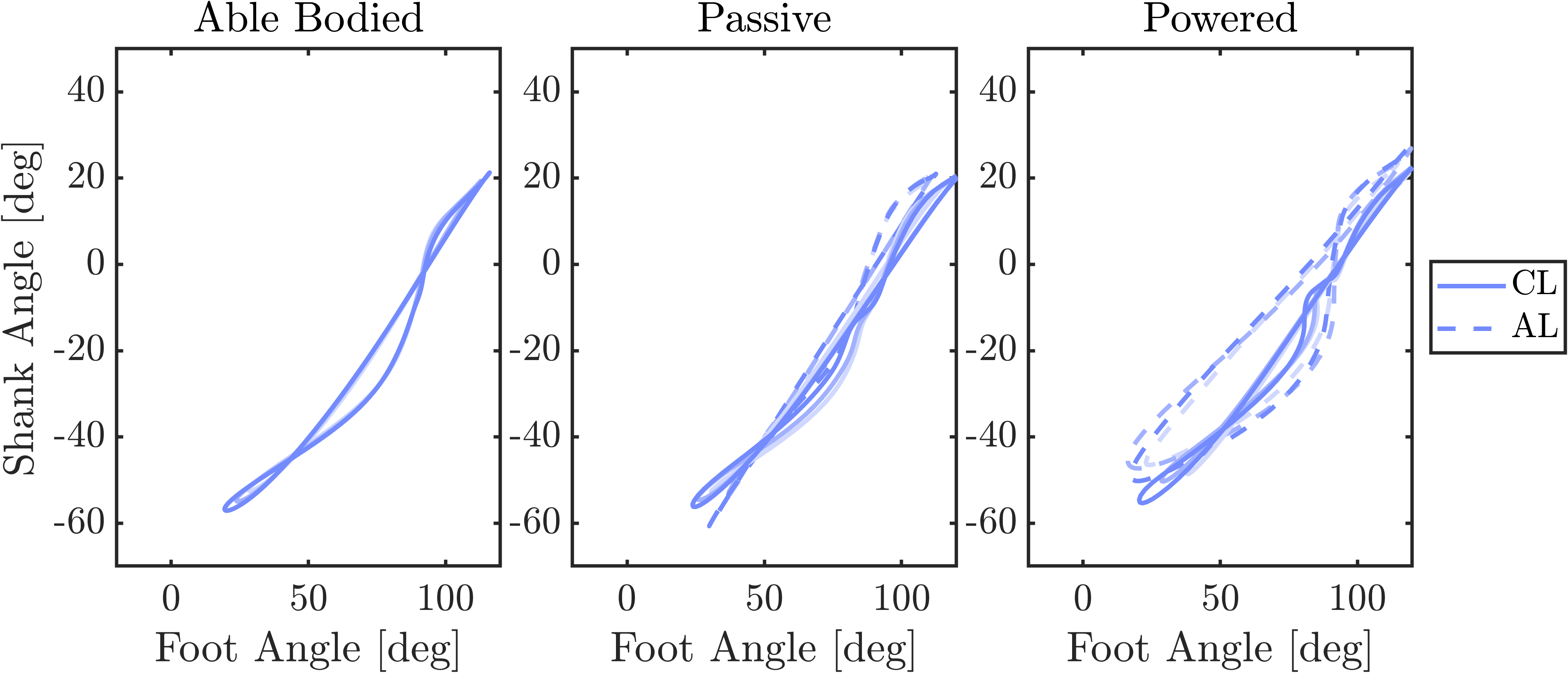}
    \caption{Mean Shank-Foot coordination which is adopted from \cite{borghese_kinematic_1996}.}
    \label{fig:4_shank_foot}
\end{figure}

The PC scores in \cref{fig:3_elevation_pcs}, show a planar coordination with a maximal out of plane component of $10.37\degree$ for AB, $8.86\degree$ for TFA passive, and $15.1\degree$ for TFA powered. 
Additionally, PC1 shows remarkable similarity to the shank and foot elevation angles implying some level of foot-shank coupling, and PC2 shows similarity to the thigh elevation angle (\cref{fig:1_elevation_angles}). 
The shank-foot coupling is shown more explicitly in \cref{fig:4_shank_foot}, and is quantified in \Cref{tab:shank_foot_linear_fit}, where the slope and bias were assessed for the linear portion of shank-foot coupling associated with swing. The passive leg has a steeper slope and stronger negative bias of approximately $10\degree$ compared to that of the AB subjects. Meanwhile, the powered leg trended in the opposite direction, with a shallower slope and lower negative bias of over $10\degree$ relative to that of AB subjects.
% Table of shank-foot cooridionation

% 4_shank_foot

\begin{table}[h!]
\centering
\caption{Slope and Bias of the linear portions for all strides in the shank-foot coordination (mean plots shown in \cref{fig:4_shank_foot}). All fits achieved a mean $R^2$ value greater than $0.9890$ with a near perfect fit in the amputated leg of the passive case ($R^2=0.9998 \pm 0.0003$).}
\begin{tabular}{l l c c}
\toprule
\textbf{Case} & \textbf{Leg} & \textbf{Slope} & \textbf{Bias $(deg)$} \\
\midrule
Able Bodied & ---               & $0.86 \pm 0.05$ & $-80.45 \pm 3.78$  \\
\midrule
\multirow{2}{*}{Passive} & Amp. & $1.00 \pm 0.01$ & $-90.44 \pm 2.37$  \\
                         & Con. & $0.82 \pm 0.05$ & $-79.01 \pm 4.75$  \\
\multirow{2}{*}{Powered} & Amp. & $0.71 \pm 0.05$ & $-59.01 \pm 3.91$  \\
                         & Con. & $0.83 \pm 0.07$ & $-77.89 \pm 5.13$  \\
\bottomrule
\label{tab:shank_foot_linear_fit}
\end{tabular}
\end{table}
%%%%%%%%%%%%%%%%%%%%%%%%%%%5
% Planarity Index
\subsection{Planarity Index (PI)}
\begin{figure}[h!]
    \centering
    \includegraphics[width=1\linewidth]{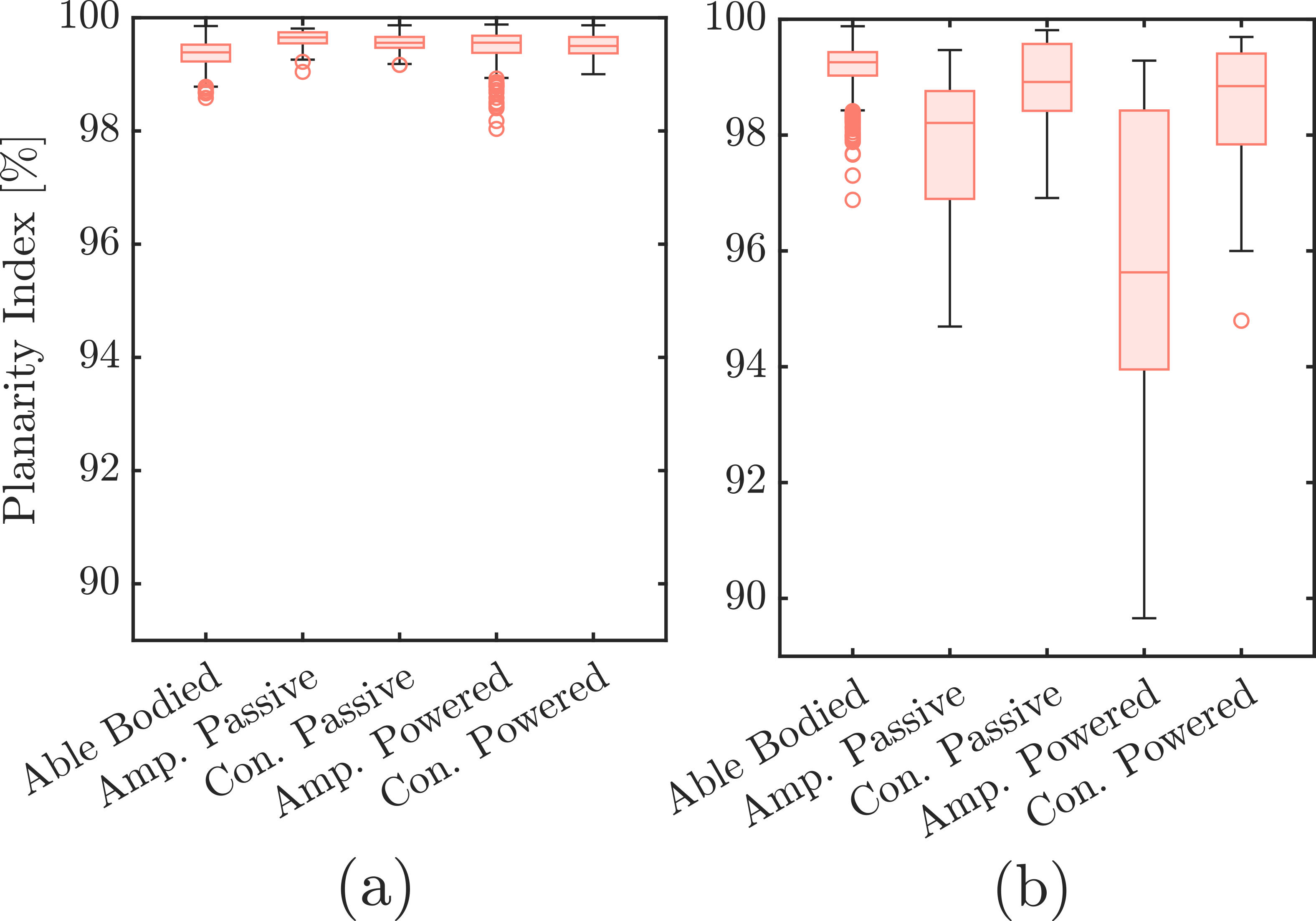}
    \caption{Planarity Index for Able Bodied, Contralateral (Con.) and Amputated (Amp.) cases for (a) elevation angles and (b) elevation space moments.}
    \label{fig:PIs_angles_moments}
\end{figure}

The elevation angle Planarity Index is presented in \cref{fig:PIs_angles_moments}a. Across all conditions the mean planarity index was $>99\%$, showing a high level of planar covariation. The ESM Planarity Index only showed high mean planarity values ($>98\%$) for the AB and contralateral legs; however, the amputated legs showed greater variance and lower mean planarity scores of $97.8 \pm 1.3[\%]$ and $95.9 \pm 2.4[\%]$ for the passive and powered legs, respectively. 

% 3_elevation_pcs

\begin{figure}[thb]
    \centering
    \includegraphics[width=1\linewidth]{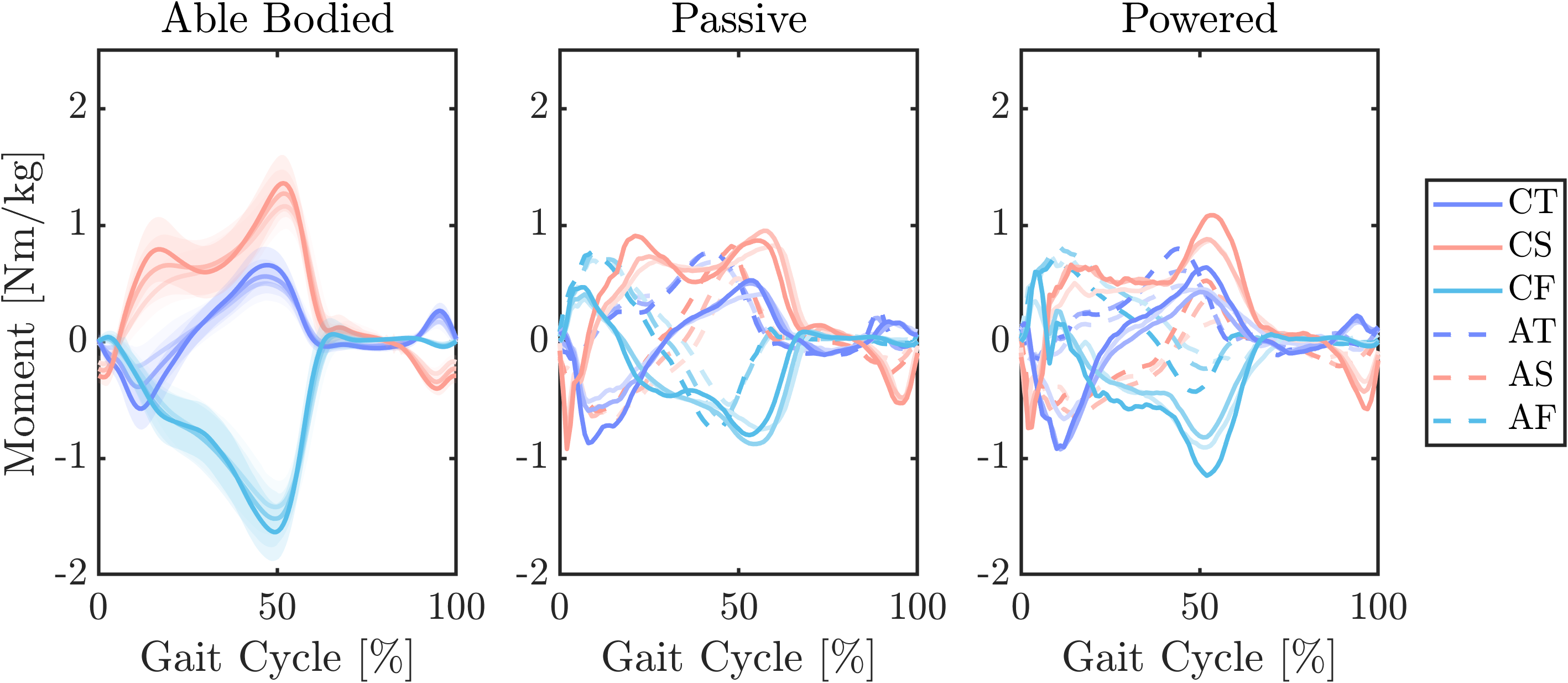}
    \caption{Elevation space moments computed from \eqref{eq:transformedMoments}.}
    \label{fig:6_elevation_moments}
\end{figure}
\begin{figure}[thb]
    \centering
    \includegraphics[width=1\linewidth]{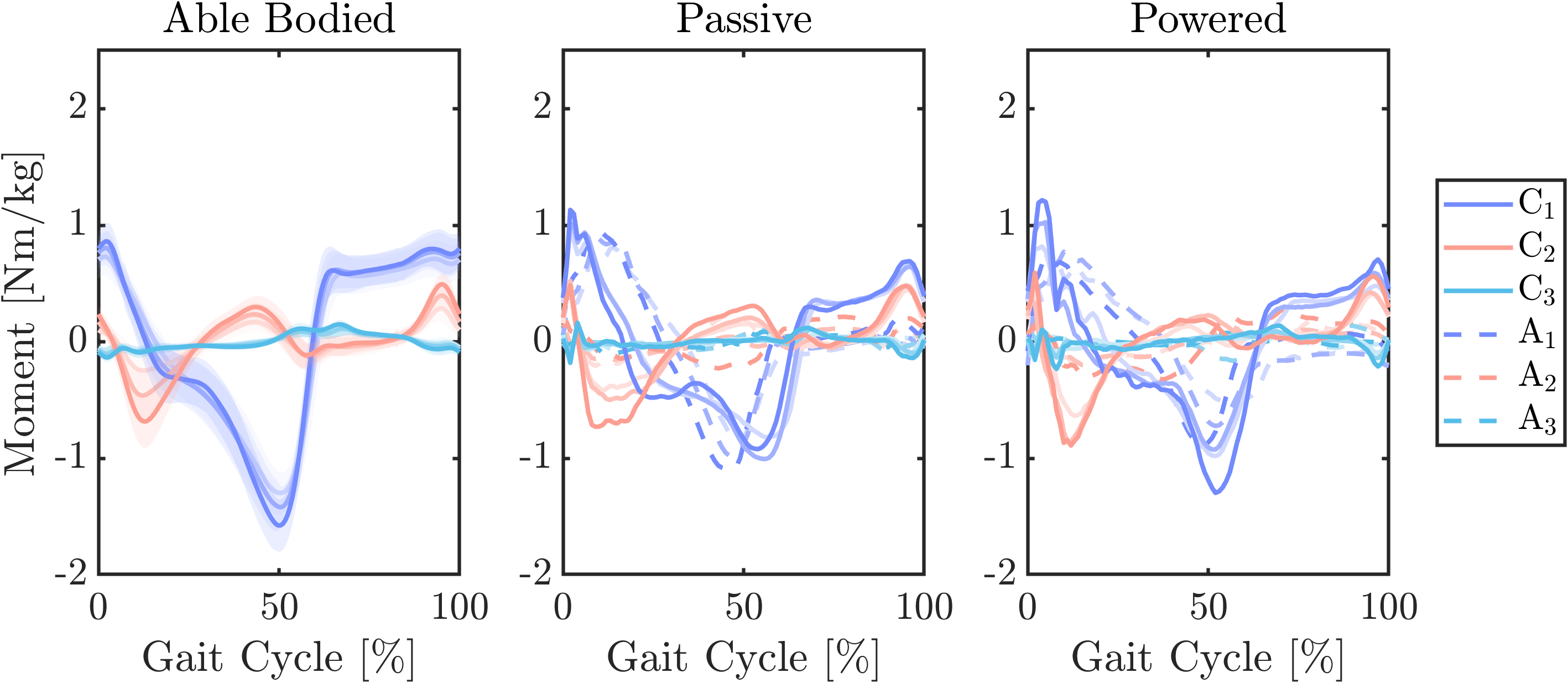}
    \caption{Principal Component scores of the elevation space moments shown in \cref{fig:6_elevation_moments}.}
    \label{fig:7_elevation_moments_pcs}
\end{figure}
\subsection{Elevation Space Moments (ESM)}
The ESM are shown in \cref{fig:6_elevation_moments} for AB and TFA subjects across speeds. The elevation moments are strikingly altered between TFA and AB conditions. Particularly, the shank and foot moments are decreased for both the contralateral and amputated legs.
The PCs of the ESM (see \cref{fig:7_elevation_moments_pcs}) show a similar coupling between the shank and foot as previously described for elevation angles. Importantly, the amputee results demonstrate altered PC scores relative to AB subjects, with large differences between amputated and contralateral legs, reflecting changes to the shank and foot moments as described in \cref{fig:6_elevation_moments}.
However, PC2 for the contralateral leg of TFA subjects is similar to AB subjects.

\subsection{Power Evaluation}\label{sec:powerEvaluation}
\begin{figure}[thpb]
    \centering
    \includegraphics[width=0.9\linewidth]{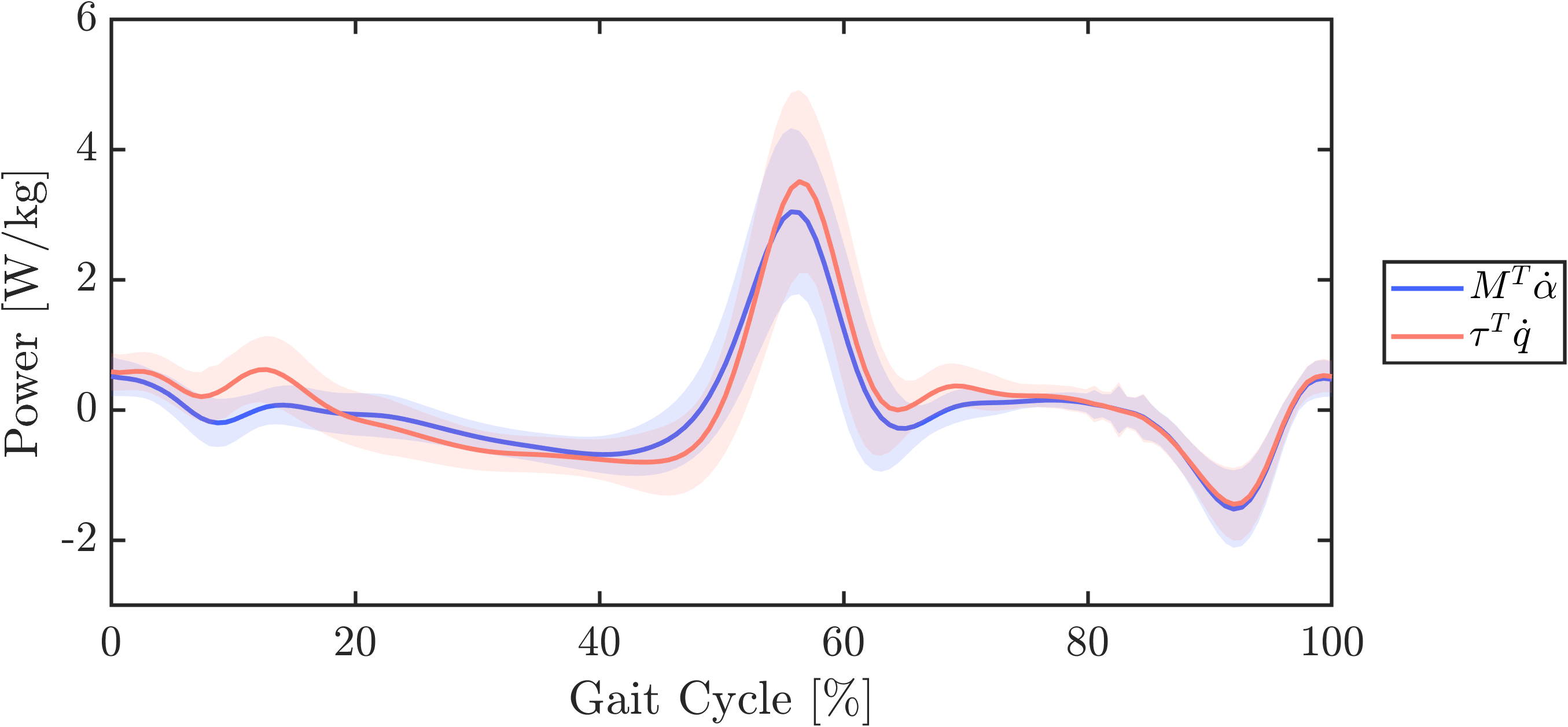}
    \caption{Means across AB subjects of the sum of internal joint powers in all 3 anatomical axes (red) compared with the sum of powers computed using the sagittal plane ESMs of the thigh, shank and foot selected from \eqref{eq:transformedMoments}, and the corresponding elevation angle velocities from \eqref{eq:alphaDot} (blue).}
    \label{fig:PowerEvaluation}
\end{figure}

The total mechanical power was computed for AB individuals both in joint space ($P_{T}=\tau^T\dot q$) and in sagittal plane elevation space ($M_\alpha ^T\dot \alpha$) and presented in \cref{fig:PowerEvaluation}. Agreement was quantified using the coefficient of determination $R^2=0.91 \pm 0.07$ and a range-normalized RMSE of $6.5\% \pm 2.2\%$. 

\subsection{Predicted Shank Profile}\label{sec:predictShankProfile}
\begin{figure*}
    \centering
    \includegraphics[width=1\linewidth]{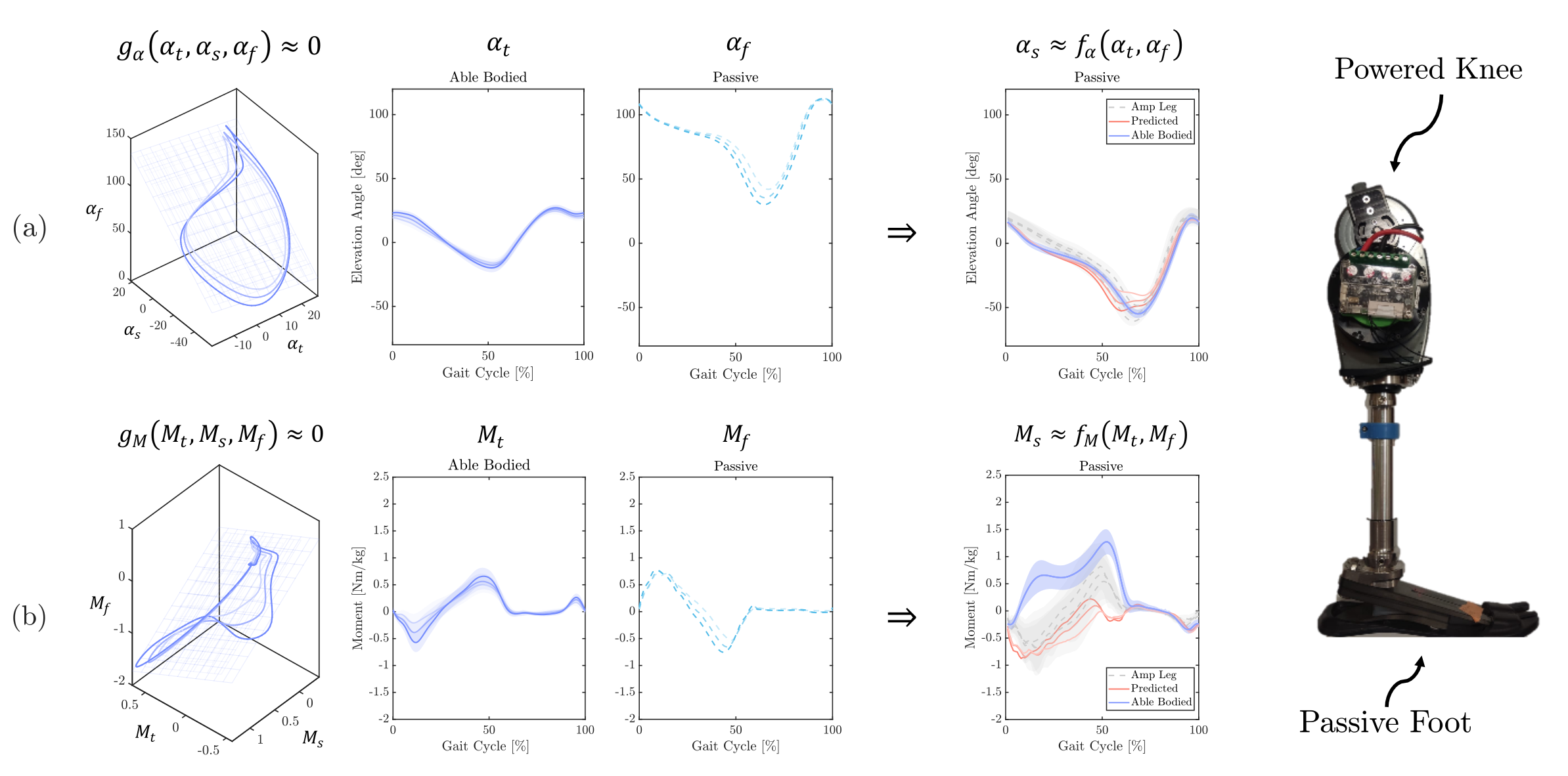}
    \caption{Demonstration of using the CVP as a constraint shown on left for (a) elevation angles, and (b) elevation space moments to reduce the dimensionality of the coordination problem in control. For each variable (angle/moment) we attempt to mimic healthy thigh ($\alpha_t, M_t$) and predict the shank behavior ($\alpha_s, M_s$) that would be required to compensate for the passive prosthetic foot ($\alpha_f, M_f$) and that would comply with healthy CVP. The AB shank (blue solid line) and amputated Passive shank (grey dashed line) angle/moments are shown for comparison on the right.}
    \label{fig:mainFigure_Demonstration}
\end{figure*}

The results for our predicted shank elevation angle profile based on the CVP constraint are shown in \cref{fig:mainFigure_Demonstration}a. Comparing the predicted shank elevation angle to the AB and TFA (Passive) profiles, reveals that to adhere to the CVP of AB subjects, for much of the gait cycle, the shank should be moving with a similar elevation angle to that of AB subjects. However, during terminal-stance/early swing, the shank angle magnitude should be reduced to compensate for the altered passive foot angle. 
The predicted shank ESM are shown in \cref{fig:mainFigure_Demonstration}b. When constrained to comply with a planar law, our predicted shank moment is found to be lower than that currently achieved by the amputated leg during terminal stance with the passive prosthetic foot.

%%%%%%%%%%%%%%%%%%%%%%%%%%%%%%%%%%%%%%%%%%%%%%%%%%%%%%%%%%%%%%%%%%%%%%%%%%%%%
\section{Discussion and Conclusion}

% Kinematic results (Elevation Angles) for AB vs. TFA subjects; high dimensionality reduction (PI > 99%) in both. First time ISC for TFA with powered devices.
In this work, we introduced a method (see \cref{sec:joint2elevation}) to compute elevation angles from 3D anatomical joint angles to make intersegmental coordination analysis more accessible. Though reporting joint angles has become standard for the evaluation of human movement, angles/moments in elevation coordinates have been shown here to reveal attributes of human motion that are not evident when considering angles/moments in joint coordinates alone. This provides motivation to the biomechanics community to report empirical gait data in elevation coordinates when possible. Our ISC3d toolbox was developed explicitly with the goal to facilitate the use of elevation angles for such comparisons.

In this work we evaluate coordination of gait for AB and TFA subjects.
\Cref{fig:1_elevation_angles,fig:2_elevation_cvp,fig:3_elevation_pcs} show that covariation of elevation angles occurs in amputee gait, but it is altered with respect to AB subjects. To our knowledge, this is the first ISC analysis of amputees walking with powered devices. We evaluated Planarity Index for elevation angles and found high mean values of $>99\%$ showing clear dimensionality reduction (see \cref{fig:PIs_angles_moments}a).

% Shank-foot coordination differences in TFA legs (rigid behavior in passive legs, coordination still altered in powered legs)
This dimensionality reduction, as shown by PI, is largely due to the coordination of the shank and foot (see \cref{fig:4_shank_foot}). Comparing the shank-foot coordination of the amputated and contralateral legs, we note a distinct difference: while the contralateral leg has a slope and bias close to that of the able-bodied individuals, the amputated leg has a clear offset. For the passive leg, the slope of the linear portion of the shank-foot coordination for the amputated leg was $1.00 \pm 0.01$ as shown in \Cref{tab:shank_foot_linear_fit}, implying that the shank and foot moved as rigid bodies, with no ankle flexion during the swing phase. Moreover, for the powered leg the results correspond to the decreased peak shank elevation angle and increased peak foot elevation angle during walking, as shown in \cref{fig:1_elevation_angles}. Thus the addition of power in this case is insufficient for resolving the coordination of the amputated leg during swing.

% Transforming moments from joint-space to elevation-space, high ESM coordination results for AB subjects.

The transformation from joint angles to elevation angles further enabled the transformation of joint moments to elevation space moments (ESM), found in the same subspace where coordination is achieved. This extension of ISC from kinematics to dynamics has not been demonstrated previously, to our knowledge. Here we present this novel method (see \cref{sec:elevJacobian}) which is freely available for use in our ISC3d toolbox \cite{siman_tov_isc3d_2026}. Evaluating the elevation space moment results in \cref{fig:6_elevation_moments}, we find a surprisingly high level of coordination between the moments in this elevation space for AB gait. This is shown in \cref{fig:PIs_angles_moments}b, where the mean PI of elevation space moments for AB gait was found to be $>99\%$ and further highlighted in the PC scores as shown in \cref{fig:7_elevation_moments_pcs} where the PCs show distinct dimensionality reduction with little variance accounted for by PC3 for AB subjects. 

% ESM coordination results for TFA subjects, lack of coordination and high variance on the amputated side.
Unlike the TFA elevation angles, which were altered yet remained coordinated, the TFA ESMs in \cref{fig:6_elevation_moments} for both passive and powered legs showed a lack of coordination which differed greatly from that of AB subjects. This was further highlighted in the results of the ESM Planarity Index (see \cref{fig:PIs_angles_moments}b). Particularly of note is that while the contralateral leg achieved fairly similar PI values as compared with AB subjects, the amputated leg had substantially decreased planarity, and higher standard deviations versus AB and contralateral legs (3.4x and 6.8x as high for the passive and powered legs, respectively). 

% Validation of the  Jacobian method ($J_\alpha$) via reconstructed powers (6.5% RMSE) and motivation for a full-order Jacobian
We evaluated our transformation ($J$) and the reduced variables for the thigh, shank and foot in the sagittal plane by comparing the reconstructed powers in \cref{sec:powerEvaluation}. The powers for AB subjects were reconstructed by the 3-DOF ESM with a relative RMSE of $6.5\% \pm 2.2\%$. This difference arises from neglecting frontal and transverse plane power terms, which may in part also explain the lower PI in the amputated legs of TFA subjects compared to AB subjects in ESM (see \Cref{fig:PIs_angles_moments}). This provides motivation to expand our analysis of ISC to 3D (such as including the frontal plane), both in elevation angles and in elevation space moments.

In future work we aim to address several limitations of this current work. The size of the dataset analyzed was fairly small, particularly for TFA subjects with only three individuals walking on their prescribed passive leg and on a powered leg. Moreover, we only considered walking conditions on a treadmill with no incline and a small range of speeds. In the future, we aim to expand our analysis to include a larger number of subjects and conditions. Additionally, we plan to test the effect of perturbing ISC and we will also include assessments of metabolic cost of transport to further validate the relationship between ISC and energetics.

% High-level impact: Extension of ISC to dynamics, potential for a unified theory of dynamic coordination, and implications for prosthetic control.
This work presents a promising direction towards improving control of powered prosthetic legs. 
Our hypothesis is that altered coordination between the residual limb and prosthesis segments is the cause of inefficient gait and elevated energy expenditure.
Our results indeed show that amputee gait exhibits altered coordination with current prosthetic systems. 
Though existing prosthetic controllers parameterize gait phase variables thus implicitly imposing some coupling between degrees of freedom; however, to our knowledge there is still no well-established explicit coordination-driven control method, relating between joints in a given prosthetic leg. Explicitly coupling joints of a knee-ankle prosthesis, with ISC as a constraint, might achieve reduced hip compensation. 
 
% Our proposed heuristic: Using AB coordination constraints to predict shank trajectories and reduce hip compensation.
We suggest an ISC-driven control method to explicitly control for coordination using a powered knee when attached with a passive ankle-foot \cref{fig:mainFigure_Demonstration}.
Specifically, we will impose a constraint in elevation angle space (ISC), while actively controlling the knee in joint space using our transformations.
Since our goal would be to achieve healthy thigh angle/moment the desired profile for thigh angle/moment can be found using a continuous phase estimator, while the foot angle/moment with a passive prosthetic foot can be measured online using an IMU and a loadcell. The desired shank angle/moment can then be computed using the combination of the thigh and foot angles/moments with healthy coordination (ISC). 
Since coordination was identified both in angle and moment space, this provides two distinct ISC-driven mid-level control frameworks, namely torque control and position control (e.g., during stance and swing, respectively). 
Using ISC, such as in this example, we can shift the focus in the field from mimicking healthy gait at a single joint level, toward mimicking healthy \textit{coordination} between the residual limb and the prosthesis. 

% for a TFA with a passive prosthetic foot, we can use healthy ISC to compute the desired shank angle/moment. 
% that would compensate to coordinate with the thigh and foot segments b ased on the healthy AB constraint. 

% Our rationale is that by folloiwng ISC, we may reduce compensatory movements in the thigh because it is what the user would expect naturaly.
% aim to identify the necessary shank angle/moment that would compensate to coordinate with the thigh and foot segments b ased on the healthy AB constraint.

%% Big Picture why do we care about this work, what is the impact?

% Future work focus on 3D dynamics, energetic costs (metabolic/mechanical energy), and the causal links between coordination and energetics

We believe that extending ISC from kinematics to dynamic coordination could potentially help address the relationship between coordination and energetics. Previously, the results from \cite{luigi_bianchi_kinematic_1998} showed a strong correlation between mechanical energy cost and the CVP, and experiments using our control framework could help address this fundamental question. The causality of this correlation has not yet been demonstrated, so in future work we aim to assess causality by perturbing coordination and measuring the resultant changes to energetics. Ultimately, the goal is to develop a means to reduce metabolic cost in amputees based on a deeper understanding of dynamic coordination and its ties to energetics. 

We present results showing dimensionality reduction in moment space. This novel approach suggests that ISC \textit{extends} beyond kinematic variables, which we believe could help elucidate open questions within the field of neural control of movement, such as the origin of ISC. Intersegmental coordination may represent an objective function of the Central Nervous System (CNS) for reducing the dimensionality of gait, and it remains an open area of debate whether the coordination is a controlled variable in human motor control \cite{barliya_analytical_2009,ivanenko_modular_2007}, or merely a kinematic result of a mechanical performance metric \cite{hicheur_intersegmental_2006}. Furthermore, intersegmental coordination may tie to the open question of whether the human nervous system is controlled in joint or global coordinates \cite{grasso_motor_1998,barliya_human_2024}. By extending ISC from kinematics to dynamics we present what may be the first result towards a unified theory of dynamic coordination.

% This is why the extension of ISC from kinematics to dynamics has not been performed previously, to our knowledge.

% Limitations (small TFA sample size, treadmill-only conditions) and mention of future expansion plans

%%%%%%%%%%%%%%%%%%%%%%%%%%%%%%%%%%%%%%%%%%%%%%%%%%%%%%%%%%%%%%%%%%%%%%%%%%%%%
%% APPENDIX %%%%%%%%%%%%%%%%%%%%
% \clearpage
\appendix
% \subsection{Definitions}\label{appendix:defVariables}
\subsection{Sign Conventions and Definitions}\label{appendix:signCorrections}

\Cref{eq:pelvisRotMatrix,eq:thighRotMatrix,eq:shankRotMatrix,eq:footRotMatrix} require sign corrections (see \Cref{tab:sign_conv}) depending on laterality and anatomical angle definitions. 

\begin{table}[thpb]
    \centering
    \caption{Sign corrections for joint rotations based on laterality.}
    \label{tab:sign_conv}
    \begin{tabular}{llcc}
        \toprule
        \textbf{Joint} & \textbf{Component} & \textbf{Left ($L$)} & \textbf{Right ($R$)} \\
        \midrule
        \textbf{Pelvis} & Tilt & $R_z(-\phi_p)$ & $R_z(-\phi_p)$ \\
                        & Obliquity & $R_x(+\delta_p)$ & $R_x(-\delta_p)$ \\
                        & Rotation & $R_y(+\rho_p)$ & $R_y(+\rho_p)$ \\
        \midrule
        \textbf{Hip}    & Flexion & $R_z(+\phi_h)$ & $R_z(+\phi_h)$ \\
                        & Adduction & $R_x(-\delta_h)$ & $R_x(+\delta_h)$ \\
                        & Rotation & $R_y(-\rho_h)$ & $R_y(+\rho_h)$ \\
        \midrule
        \textbf{Knee}   & Flexion & $R_z(-\phi_k)$ & $R_z(-\phi_k)$ \\
                        & Adduction/Varus & $R_x(-\delta_k)$ & $R_x(+\delta_k)$ \\
                        & Rotation & $R_y(-\rho_k)$ & $R_y(+\rho_k)$ \\
        \midrule
        \textbf{Ankle}  & Dorsiflexion & $R_z(\phi_a + 90^\circ)$ & $R_z(\phi_a + 90^\circ)$ \\
                        & Rotation & $R_x(-\rho_a)$ & $R_x(+\rho_a)$ \\
                        & Inversion & $R_y(+\delta_a)$ & $R_y(-\delta_a)$ \\
        \bottomrule
    \end{tabular}
\end{table}

This is further explained in our ISC3d toolbox \cite{siman_tov_isc3d_2026}. Joint angles follow Vicon Plug-In Gait definitions of anatomical angles \cite{vicon_plug_in_gait}. Reference frames for transformations follow ISB recommendation \cite{wu_isb_1995}.
For the $j^{th}$ joint, we denote anatomical angles as flexion ($\phi_j$), adduction ($\delta_j$) and rotation ($\rho_j$). The joint angles vector ($q$) in \eqref{eq:alphasDotEQJqdot}, is defined in \eqref{appendixeq:q}. 
\begin{equation}\label{appendixeq:q}
    q = [\phi_p,\delta_p,\rho_p,\phi_h,\delta_h,\rho_h,\phi_k,\delta_k,\rho_k,\phi_a,\delta_a,\rho_a]
\end{equation}
The joint torque vector ($\tau$) used in \eqref{eq:transformedMoments}, is defined in \eqref{appendixeq:tau}. It contains hip ($\tau_h$), knee ($\tau_k$), and ankle ($\tau_a$) torques in three axes $\tau_j=[\tau_{\phi_j},\tau_{\delta_j},\tau_{\rho_j}]$, augmented with zeros to match the dimensions of $q\in \mathbb{R}^{12}$ and $J\in \mathbb{R}^{12\times 12}$. Note that setting $\tau_p=0_{1\times3}$ is arbitrary, because the resulting terms for $M_t, M_s,M_f$ are independent of $\tau_p$.
\begin{equation}\label{appendixeq:tau}
    \tau = [0_{1\times3},\tau_h,\tau_k,\tau_a]^T
\end{equation}

% \begin{equation}
%     M\triangleq\left(\frac{\partial q}{\partial \alpha }\right )^T\tau
% \end{equation}

%%%%%%%%%%%%%%%%%%%%%%%%%%%%%%%%%%%%%%%%%%%%%%%%%%%%%%%%%%%%%%%%%%%%%%%%%%%%%
\bibliographystyle{IEEEtran}
\bibliography{sample.bib}

%%%%%%%%%%%%%%%%%%%%%%%%%%%%%%%%%%%%%%%%%%%%%%%%%%%%%%%%%%%%%%%%%%%%%%%%%%%%%

%%%%%%%%%%%%%%%%%%%%%%%%%%%%%%%%%%%%%%%%%%%%%%%%%%%%%%%%%%%%%%%%%%%%%%%%%%%%%

\end{document}